\documentclass{article}
\usepackage{ijcai23}

\usepackage[natbib=true,style=authoryear]{biblatex}
\usepackage{times}
\usepackage{soul}
\usepackage{url}
\usepackage[hidelinks]{hyperref}
\usepackage[utf8]{inputenc}
\usepackage[small]{caption}
\usepackage{graphicx}
\usepackage{amsmath}
\usepackage{amsthm}
\usepackage{booktabs}
\usepackage{algorithm}
\usepackage{algorithmic}
\usepackage[switch]{lineno}

\usepackage[australian]{babel}  
\usepackage{booktabs}
\usepackage{bigstrut}
\usepackage{colortbl}

\title{
Leveraging Cutting Edge Deep Learning Based Image Matching \\
for Reconstructing a Large Scene from Sparse Images
}

\author{
Georg Bökman$^1$
\And
Johan Edstedt$^2$
\affiliations
$^1$Chalmers University of Technology
\\
$^2$Linköping University
}

\begin{document}

\maketitle

\begin{abstract}
    We present the top ranked solution for the 
    AISG–SLA Visual Localisation Challenge benchmark 
    (IJCAI 2023),
    where the task is to estimate relative motion between
    images taken in sequence by a camera mounted on a car
    driving through an urban scene.
    
    For matching images we use our recent deep learning based
    matcher RoMa.
    Matching image pairs sequentially and estimating relative
    motion from point correspondences sampled by RoMa
    already gives very competitive results -- third rank
    on the challenge benchmark.

    To improve the estimations we extract keypoints
    in the images, match them using RoMa, and perform
    structure from motion reconstruction using COLMAP.
    We choose our recent DeDoDe keypoints for their high repeatability.
    Further, we address time jumps in the image sequence by matching specific non-consecutive
    image pairs based on image retrieval with DINOv2.
    These improvements yield a solution beating all competitors.

    We further present a loose upper bound on the accuracy obtainable by the image retrieval approach by also matching hand-picked non-consecutive pairs.
\end{abstract}

\begin{figure*}
    \centering
    {
        \includegraphics[height=0.14\textwidth,angle=270]{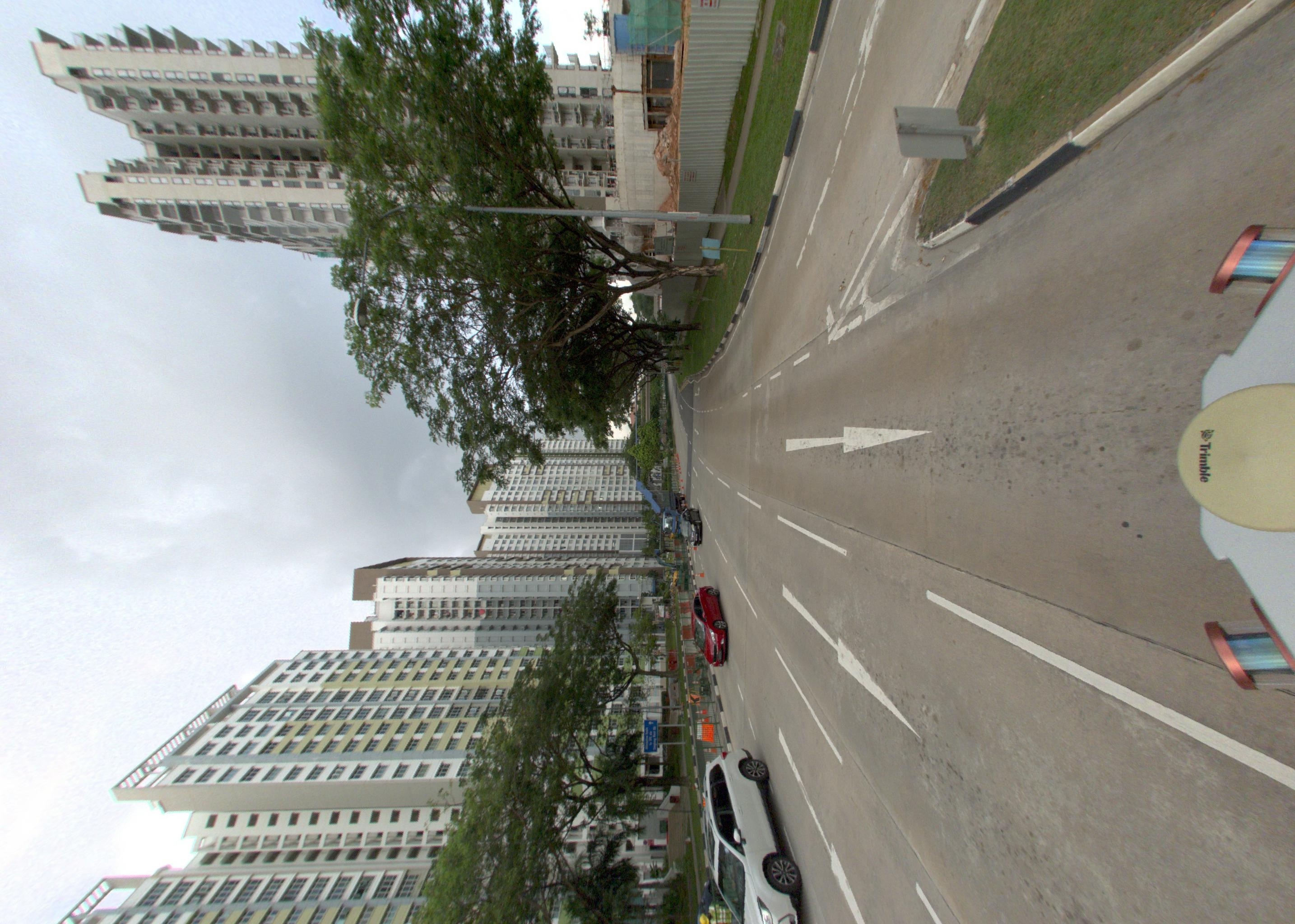}
        \includegraphics[height=0.14\textwidth,angle=270]{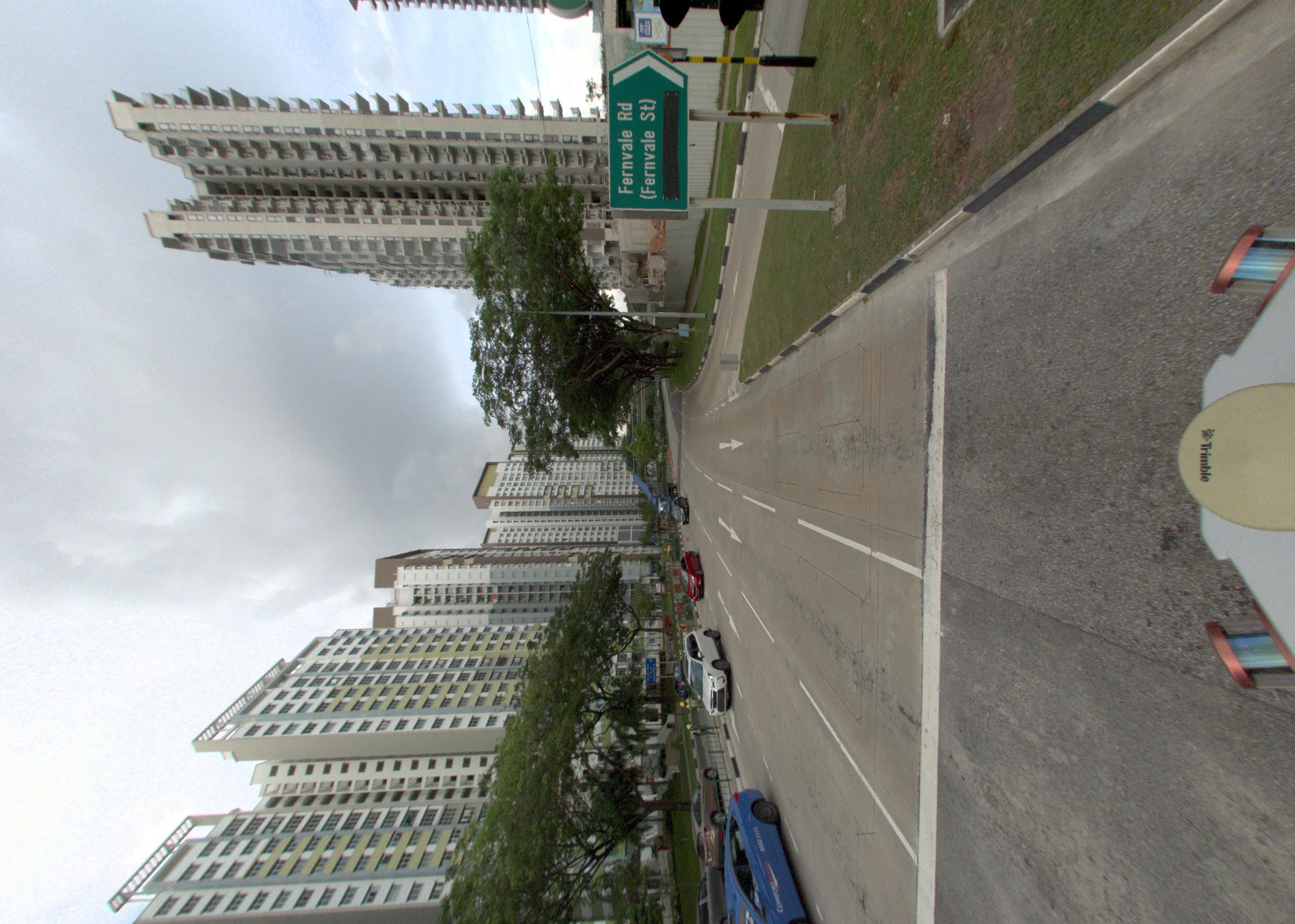}
    }
    \hfill
    {
        \includegraphics[height=0.14\textwidth,angle=270]{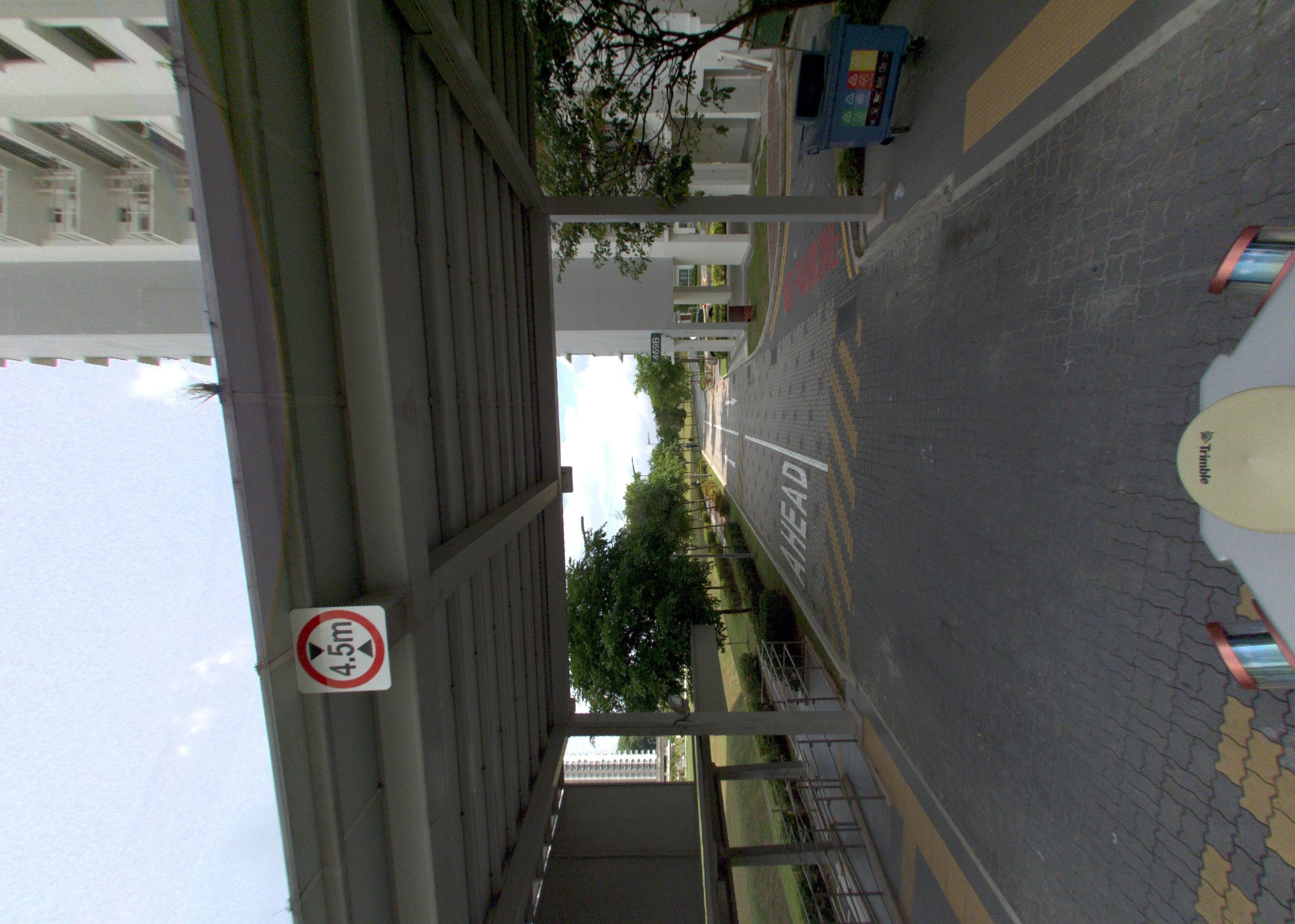}
        \includegraphics[height=0.14\textwidth,angle=270]{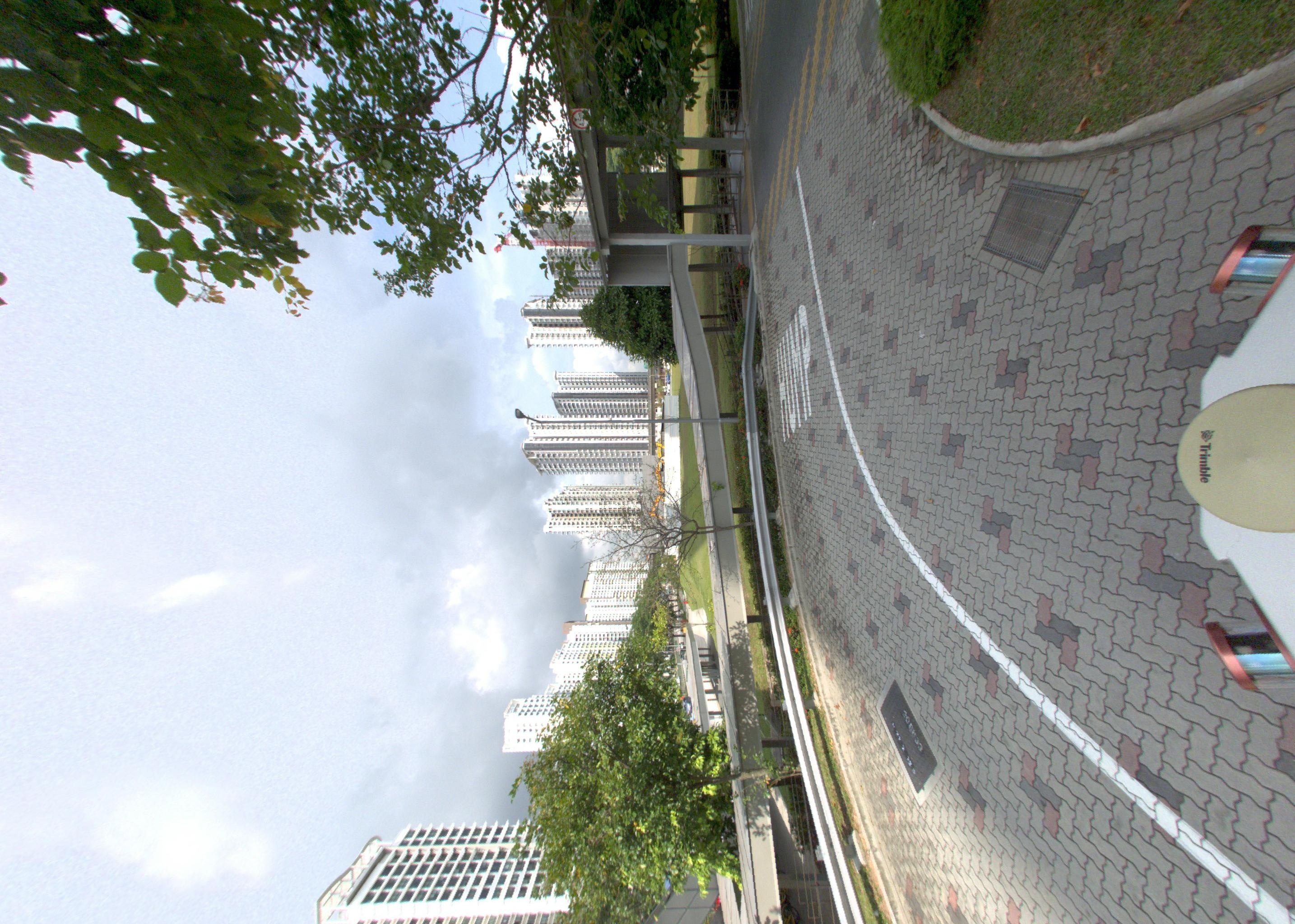}
    }
    \hfill
    {
        \includegraphics[height=0.14\textwidth,angle=270]{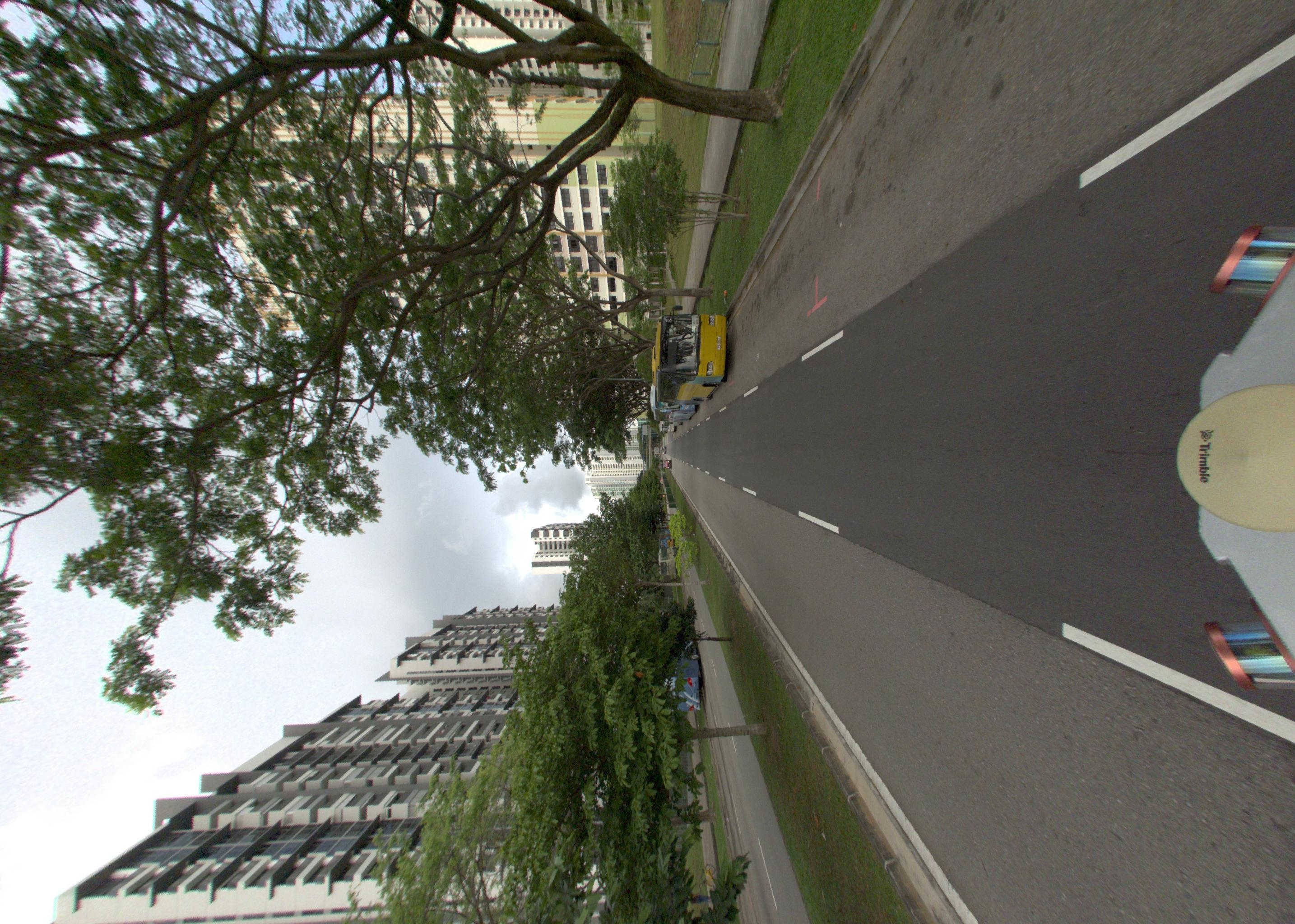}
        \includegraphics[height=0.14\textwidth,angle=270]{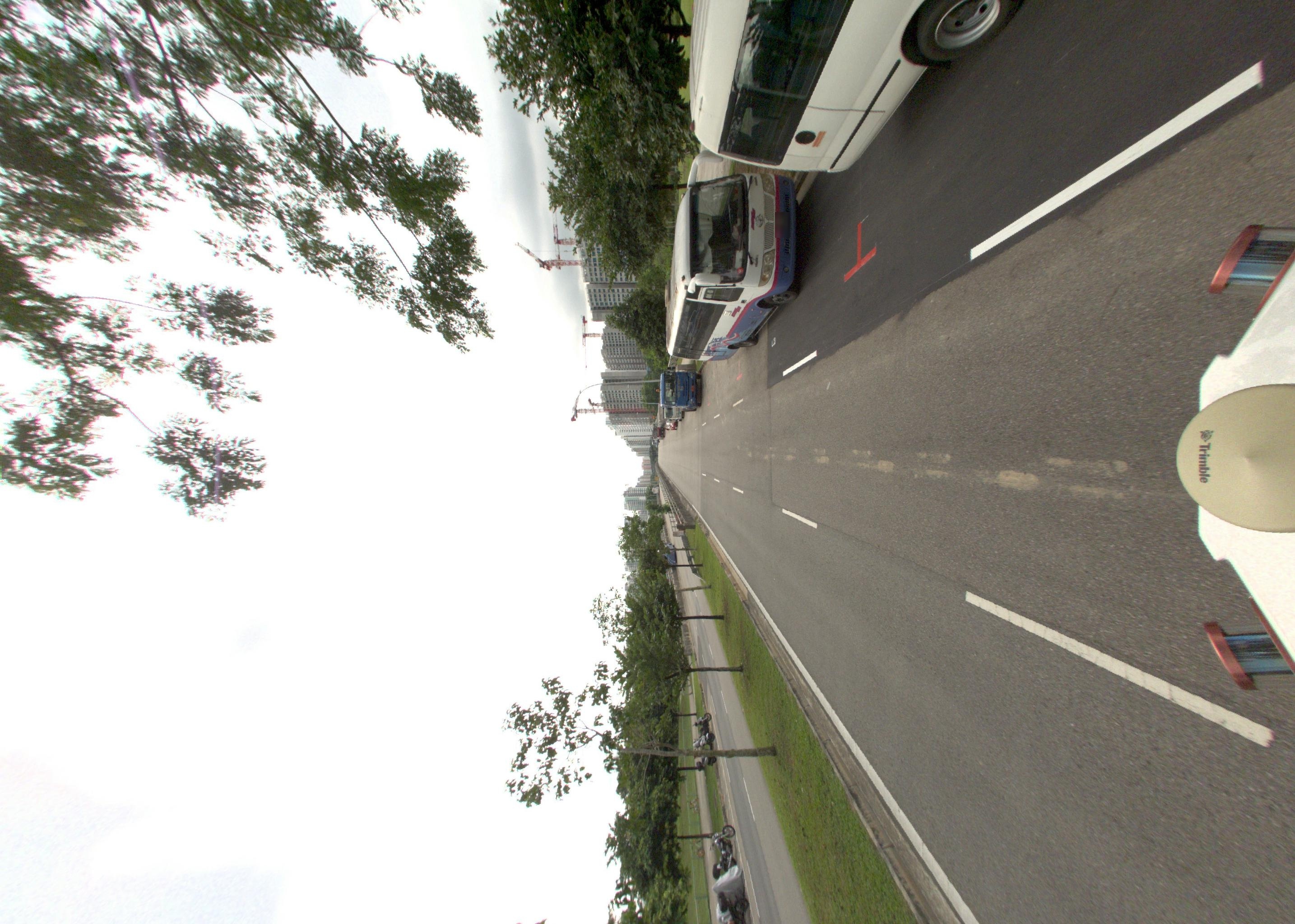}
    }
    \caption{
        Three pairs of consecutive images in the benchmark sequence.
        Note that the camera is pointing towards the back
        of the vehicle so that the movement between consecutive
        captures is out of the image.
        \textbf{Left:} A typical easy to match pair.
        \textbf{Center:} A typical difficult to match pair.
        The low frequency of captures makes sharp turns such as this very challenging.
        \textbf{Right:} A pair that is impossible to match.
        These images are consecutive in the sequence,
        but were taken 76 seconds
        apart and have no visual overlap.
    }
    \label{fig:data}
\end{figure*}

\begin{figure*}
    \centering
    \fbox{
        \includegraphics[height=0.14\textwidth]{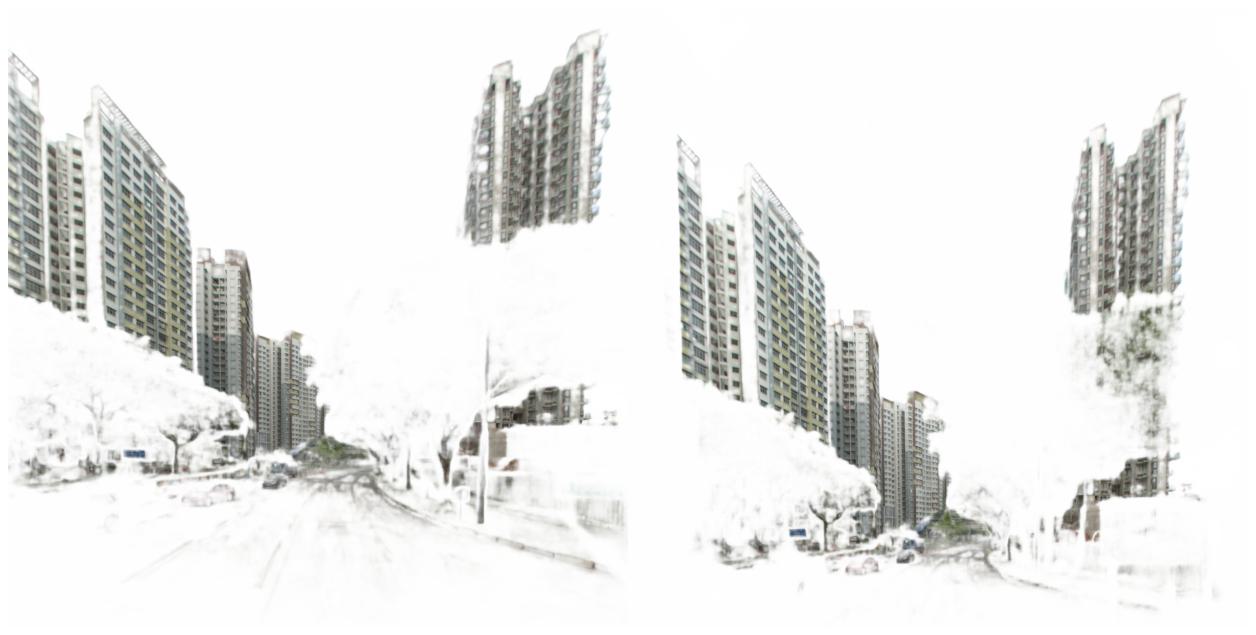}
    }
    \hfill
    \fbox{
        \includegraphics[height=0.14\textwidth]{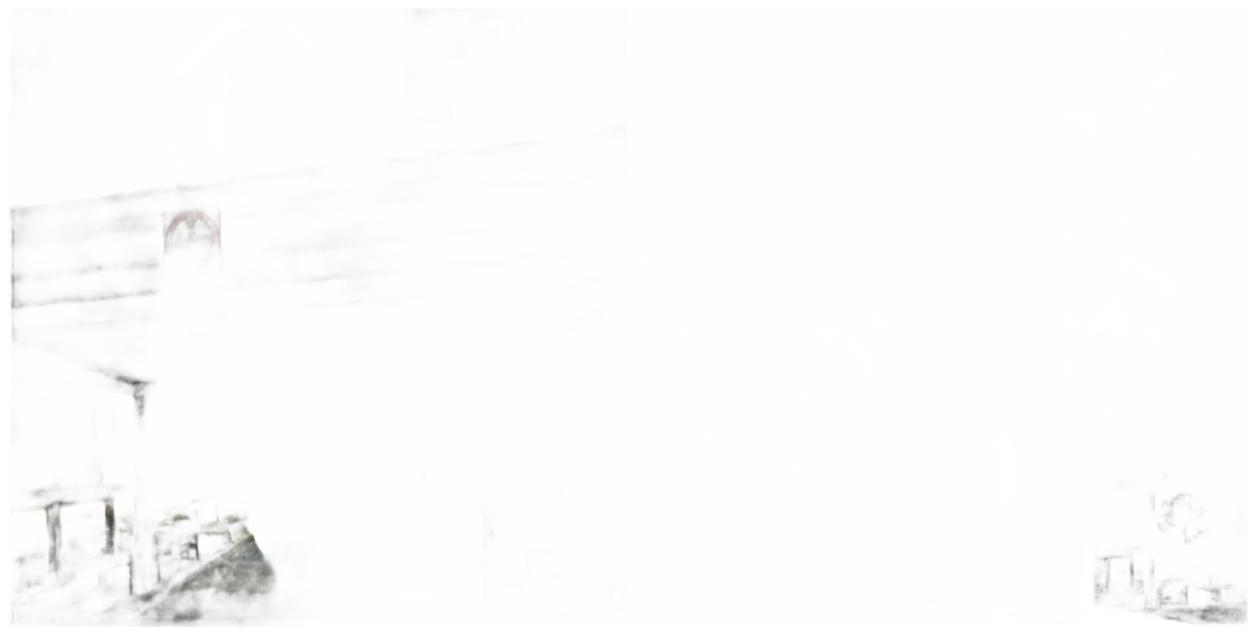}
    }
    \hfill
    \fbox{
        \includegraphics[height=0.14\textwidth]{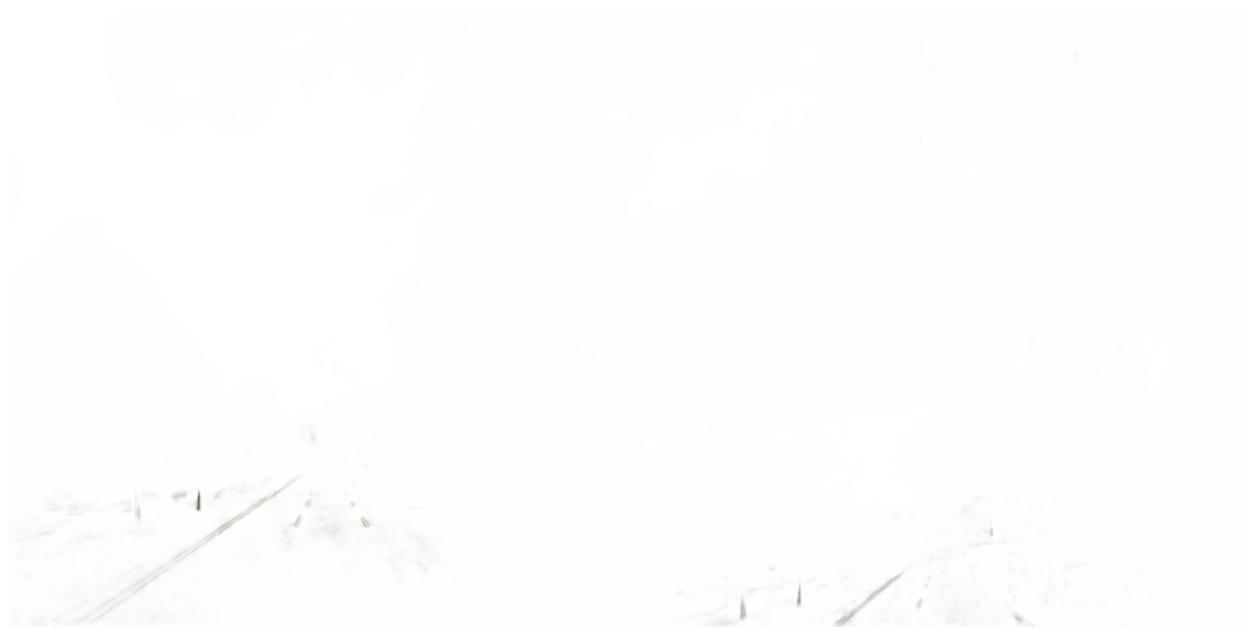}
    }
    \caption{
        Visualizations of RoMa warps for the three image
        pairs shown in Figure~\ref{fig:data}.
        The images are cropped to remove
        the visible part of the ego vehicle and 
        parts of the road before matching with RoMa.
        Each pixel is colored by the color of the location in the other image that it is warped to.
        The brighter a pixel is the less confident
        the warp is.
        RoMa matches the easy pair well and also finds
        a correct low confidence warp for the difficult pair.
        Matching for the impossible pair mostly fails,
        but the road is still matchable yielding
        an incorrect warp with very low confidence.
    }
    \label{fig:roma}
\end{figure*}

\section{Introduction}
In this technical report, we describe how we obtained the highest ranking solution on the AISG–SLA Visual Localisation Challenge~\citep{aisingaporeAISGSLAVisual} at IJCAI 2023. The challenge consists of estimating relative motion between images in a
sequence taken by a camera mounted on a car driving in an urban environment in Singapore, with the main challenge being the combination of low overlap, as well as symmetric and repeating structures. We further describe these challenges in Section~\ref{sec:data}.

The challenge hosts provide four sequences
similar to the test sequence as training data, 
but we use this data very minimally (only for estimating the scale of the scene).
Instead, we leverage our recent work on image matching through
the dense matcher RoMa~\citep{edstedtRoMaRevisitingRobust2023}
and the keypoint detector DeDoDe~\citep{edstedtDeDoDeDetectDon2023},
which are both deep learning methods trained on
the large-scale outdoor dataset MegaDepth~\citep{liMegaDepthLearningSingleView2018}.
As we do not use the challenge training data,
our method is agnostic to the specifics of the
challenge setup such as camera type and image capturing frequency.
In particular, MegaDepth does not consist
of images from a car mounted camera, which highlights
the generality of our approach.

In Section~\ref{sec:data} we discuss the challenge data
and the main insights that informed our method choices.
In Sections~\ref{sec:method1} and \ref{sec:method2} we describe
our two main methods. The first is based on simply matching
the images sequentially with RoMa and estimating relative motion
from point correspondences sampled from the obtained dense warps.
The second is more elaborate and further reconstructs the image
sequence and the observed 3D structure using the structure from motion
software COLMAP~\citep{schonbergerStructurefromMotionRevisited2016, schonbergerPixelwiseViewSelection2016}.
In Section~\ref{sec:technical} we lay out the technical details
of our solutions and in Section~\ref{sec:limitations} we discuss
limitations and ideas for the future.

\begin{table}
  \centering
  \caption{Results on AISG-SLA VLC. Ranked top to bottom by lowest rotation error.}
  \label{tab:results}%
  \footnotesize
    \begin{tabular}{lrr}
    \hline
    {} & {Rotation error} & {Translation error} \bigstrut[t]\\
    {} & [milliradians] & [meters] \bigstrut[b]\\
    \hline
    \textbf{Method II-B}   & $32$ & $1.4$ \bigstrut[t]\\
    \rowcolor[rgb]{.92,.92,.92} \textbf{Method II-A} & $34$ & $1.5$ \\
     \textbf{Method I} & $39$ & $2.0$ \bigstrut[b]\\
    \hline
    \end{tabular}%
\end{table}%

\section{Data Inspection}\label{sec:data}
We inspect the benchmark image sequences in order to
inform our methodology.
The two main difficulties of the challenge 
seem to be
i) the low overlap of consecutive image captures as 
the median time between image captures is over one second
and 
ii) large discontinuities in the sequence where 
images have no visual overlap at all.
Note that as the benchmark scores are mean
rotation error and mean 
translation error for relative motion
between consecutive images,
failing to accurately estimate the pose for even
a small fraction of pairs in the sequence
can greatly impact the score.
We exemplify the data and the two main difficulties in Figure~\ref{fig:data}.

To deal with the sparsity of captures 
and resulting low overlap between
many image pairs we use the recent state-of-the-art matcher RoMa (see Section~\ref{sec:roma}).
To deal with the discontinuities we try to stitch sub-sequences
of the image sequence by matching non-consecutive image pairs
with visual overlap.
These extra image pairs are identified either
using image retrieval with DINOv2~\citep{oquab2023dinov2}
(see Section~\ref{sec:retrieval}) or by manual
inspection of the images (see Section~\ref{sec:manual}).

Looking at the four provided training
data sequences, we observe that
two of them have a median translation around 10 meters and two
have a median translation around 5 meters.
We aim for a median translation of 10 meters
in our solutions as this gave better score than 5 meters.

\section{Method I -- Simple Sequential Matching}\label{sec:method1}
\begin{figure}
    \centering
    \fbox{
        \includegraphics[width=0.75\columnwidth]{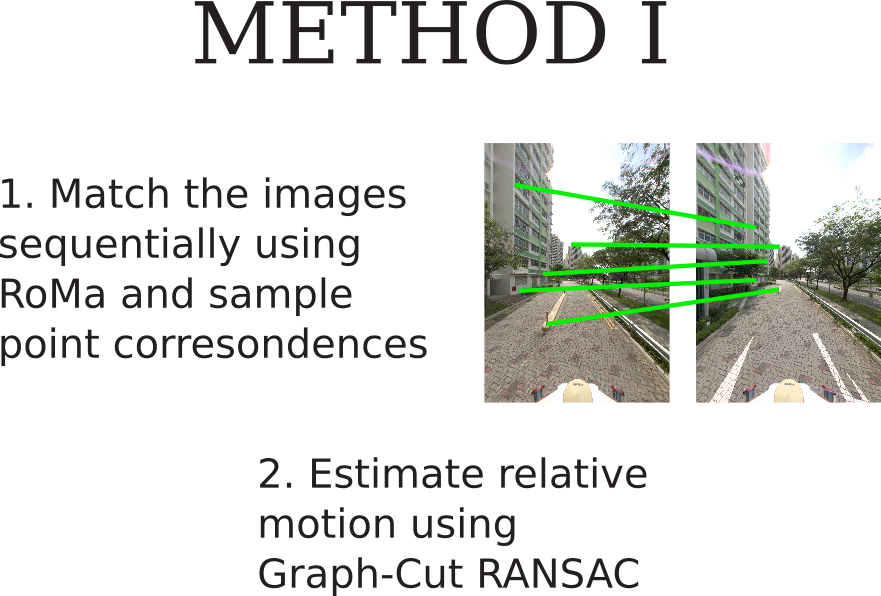}
    }
    \caption{Pipeline for Method~I.}
    \label{fig:methodI}
\end{figure}

\begin{figure*}
    \centering
    \fbox{
        \includegraphics[width=0.9\textwidth]{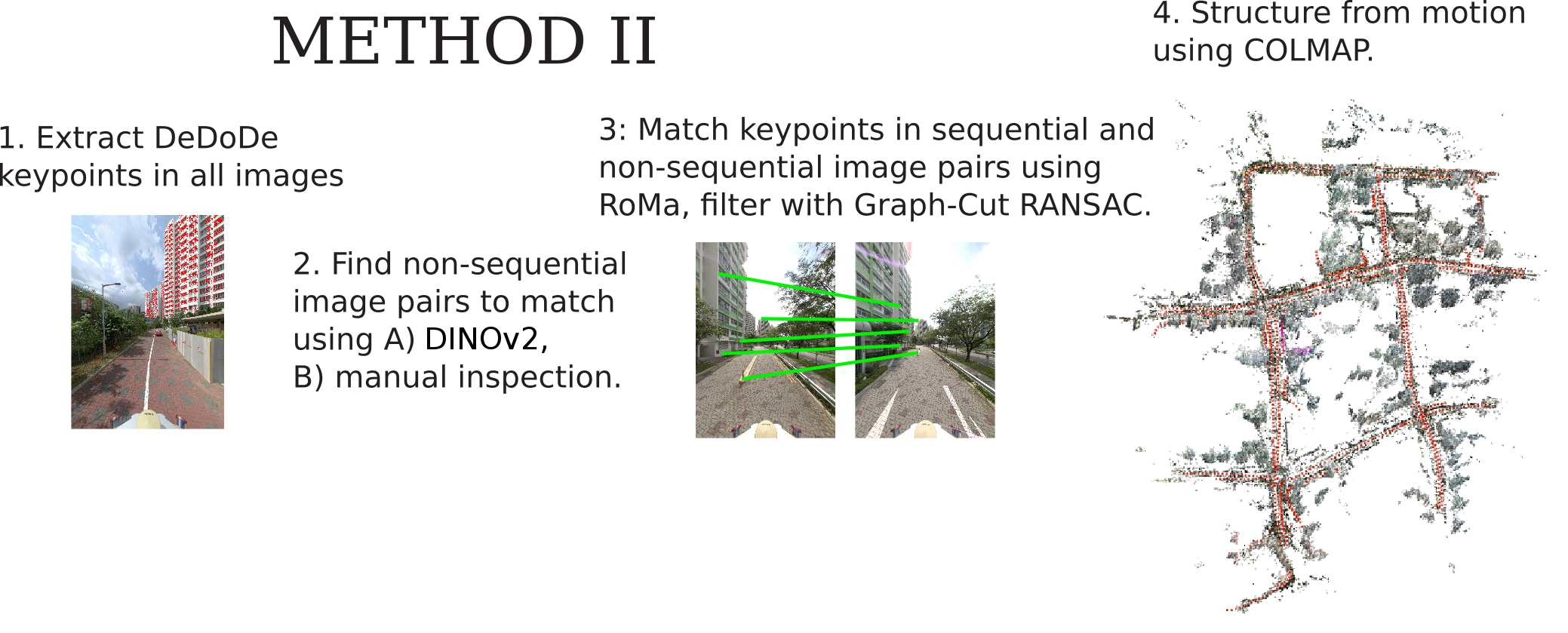}
    }
    \caption{
    Pipeline for Method~II.
    In the structure from motion visualization, each red dot is a position where a picture was taken.
    }
    \label{fig:methodII}
\end{figure*}

Our first method is based on simply matching the images sequentially using RoMa
and is illustrated in Figure~\ref{fig:methodI}.
This method, which we call Method~I, already yields very competitive results,
as seen in Table~\ref{tab:results}.
We stress that the method does not handle
the time jumps in the data, i.e.~for non-overlapping pairs the method may produce arbitrary relative pose, and could likely be
improved by a heuristic such outputting no relative motion at time jumps or when the number of inliers in estimation is below a critical threshold.
However, as we did not submit a solution with such a heuristic it is uncertain to what extent this would improve the score.
In the following subsections we will describe Method I in more detail.

\subsection{Using RoMa to Match Image Pairs}\label{sec:roma}
RoMa~\citep{edstedtRoMaRevisitingRobust2023} is our recent
dense image matcher.
It takes two images as input and outputs warps between them,
i.e. a prediction of what position
in the second image each position
in the first image corresponds to and vice versa.
RoMa also outputs confidences for the warps, so
that we know where the warps are most likely to be correct.
We visualize warps of three image pairs in Figure~\ref{fig:roma}.
We input cropped images to RoMa to avoid matches on the visible part of the ego vehicle,
which would lead to incorrect motion estimates.

\subsection{Estimating Relative Motion}\label{sec:ransac}
We sample point correspondences from RoMa for each image
pair and estimate essential matrices by using 
Graph-Cut RANSAC~\citep{barathGraphCutRANSAC2018} through
the \texttt{USAC\_ACCURATE} option in OpenCV~\citep{bradskiOpenCVLibrary2000}.
From the essential matrices we obtain relative rotations
but relative translations only up to an multiplicative factor.
We set this factor by specifying all translations
to be 10 meters long as described in 
Section~\ref{sec:data} and setting the sign so that the 
translation is facing forwards, as the car
should almost always be travelling forwards.

\section{Method II -- Matching Non-Sequential Images and Attempting to Reconstruct the Whole Scene}\label{sec:method2}
While we already achieve competitive results using Method I (see Table~\ref{tab:results}) we found, as discussed in Section~\ref{sec:data}, that it struggles at discontinuities in the sequence. At such points, the sequential estimation can fail and produce large errors.
To address this and also improve the estimation in general,
we integrate full structure from motion reconstruction using COLMAP.
We now, instead of only matching consecutive image pairs,
also add image pairs that are not consecutive in the image sequence, but that are still
depicting the same part of the scene. Some roads in the scene are in fact
driven through up to three times.
If images $a$ and $b$ are consecutive but with zero visual overlap,
we can get a good relative motion estimate from $a$ to $b$ if there
are images $c$ and $d$ with high visual overlap such that $a$ and $c$ are connected by
some sequence of image pairs with visual overlap, as are $b$ and $d$.

The new image pairs are chosen either using
image retrieval with DINOv2 (Method II-A, see Section~\ref{sec:retrieval}) or
by manually looking through the image sequence for good image pairs
(Method II-B, see Section~\ref{sec:manual}).
Both of these versions outperform all competitors on the benchmark as is seen in Table~\ref{tab:results}. Method~II is illustrated in Figure~\ref{fig:methodII}.

\subsection{DeDoDe Keypoints}\label{sec:dedode}
Our SfM software of choice, COLMAP, requires keypoints in each image and matches between them for the 3D reconstruction pipeline.
We use our very recent DeDoDe keypoints~\citep{edstedtDeDoDeDetectDon2023} which
are trained to produce consistent structure from motion tracks.
For each image pair, we use the warp generated by RoMa in order to match the DeDoDe keypoints. We use the following strategy. For each DeDode keypoint in the pair, we use bilinear sampling of the RoMa warp to produce a corresponding coordinate in the other image. If the warp certainty is lower than $0.1$ we discard the prediction. We select matches as keypoint pairs that i) are mutual nearest neighbours, and ii) the distance between the keypoint and prediction in each direction is less than $0.5\%$ of the image size. 

We further use a multicrop strategy to obtain more matches -- we match not only the original two large crops as in Section~\ref{sec:method1},
but also all combinations of the following crops: each large crop, each right half and each left half, yielding in total nine crop combinations, so nine warps that are used to match the keypoints.
We filter the matches by estimating essential matrices as in Section~\ref{sec:ransac}.

\subsection{Reconstructing the Sequence with COLMAP}\label{sec:colmap}
We use COLMAP~\citep{schonbergerPixelwiseViewSelection2016, schonbergerStructurefromMotionRevisited2016}
to perform structure from motion on the matches obtained from DeDoDe + RoMa + Graph-Cut RANSAC.
As output we get several incomplete reconstructions consisting of subsets of the
test sequence. It turns out to be very difficult to obtain a complete reconstruction due
to the low visual overlap between some consecutive images. The largest reconstructions we are able to
get contain around 1500 images out of 2219, see Figure~\ref{fig:methodII} for a demonstration of
how such a reconstruction looks -- the larger roads are more or less perfectly reconstructed while
smaller roads are missing because they contain more turns and hence more pairs with low visual overlap.

To obtain the final sequence of relative motions we go through the following steps for each consecutive image pair:
\begin{enumerate}
    \item If the pair is contained in a COLMAP reconstruction, take the relative motion from the largest reconstruction that contains it.
    \item Else, if the pair is not in a COLMAP reconstruction and has a time jump of over one minute,
    set the relative motion to zero.
    \item Else, use the relative motion estimated from the essential matrix as in Section~\ref{sec:ransac}.
\end{enumerate}
The scale of each COLMAP reconstruction is set so that it has a median translation length of 10 meters,
which corresponds to the median in the training data.

\subsection{Method II-A -- DINOv2 Image Retrieval}\label{sec:retrieval}
\textbf{Method II-A} incorporates image retrieval using DINOv2~\citep{oquab2023dinov2} to obtain visually similar image pairs
that are non-consecutive in the sequence.
We match all pairs, more than 20 captures apart,
that are within a distance of $0.35$ of each other in
DINOv2 class token embedding space, 
a threshold that was chosen as high as possible
without inclusion of too many false positives.
We found that false positives can significantly degrade
COLMAP reconstructions, leading to poor results.
Therefore, after matching the retrieved image pairs using RoMa,
we only include image matches with mean RoMa certainty
above $0.05$ in COLMAP.

\subsection{Method II-B -- Manual Search for Image Pairs}\label{sec:manual}
While adding DINOv2 image pairs yields substantial improvement, 
the retrieval is not optimal.
It is possible to improve even
further by manually adding image pairs,
which we now incorporate in order
to get a loose upper bound on how low errors 
are possible with extremely good image retrieval.
By looking at the subsequences generated by COLMAP, we identify good-to-match image pairs,
i.e. image pairs that depict the same parts of the scene but belong to disjoint subsequences.
We also specify which crops of these image pairs should be matched, which is necessary to get
good results from RoMa for some particularly tricky image pairs.
We call this final improvement \textbf{Method II-B}.
Note that it still fails frequently on turns with low/no visual overlap.
Likely, the only way to solve such cases would be use more
frequent image captures or incorporate data from another sensor.

\section{Technical Details}\label{sec:technical}
We run all experiments on a desktop computer with an AMD Ryzen Threadripper 2950X 16-Core CPU and
a single NVIDIA GeForce RTX 2080 Ti GPU with 11 GB memory.
We use Python~3.10, PyTorch~2.0 and COLMAP~3.8.

Method I runs at around 1.5 seconds per image pair, so just under one hour for the complete sequence.

Method II has a higher runtime.
Extracting keypoints takes around 10 minutes.
Then, matching is similar to Method I, but we use multiple crops and match more pairs so that the time
needed is around 10 hours (including DINOv2 image retrieval which is negligible in comparison).
Finally, running COLMAP takes a couple of hours,
the runtime
can be tweaked by setting how many initial pairs should be tried.

\section{Limitations and Untested Ideas}\label{sec:limitations}
While Method I has a low runtime, our best-performing approach, Method II, requires more compute. In particular, using COLMAP and multi-crop matching increases the runtime significantly. However, the runtime of our full approach could be reduced by i) utilizing GPU for COLMAP and ii) only using our multi-crop matching for particularly hard pairs.
We also believe that improvements on our method, both in terms of runtime and in terms of performance,
could be obtained by tuning RANSAC and COLMAP parameters, which we have not done to a large extent.

\section*{Acknowledgments}
This work was partially supported by the Wallenberg AI, Autonomous Systems and Software Program (WASP) funded by the Knut and Alice Wallenberg Foundation; and by the strategic research environment ELLIIT funded by the Swedish government.

\printbibliography

\end{document}